\documentclass[letterpaper, 10 pt, conference]{ieeeconf}  

\IEEEoverridecommandlockouts                              

\usepackage{mathtools}
\usepackage{graphics} 
\usepackage{epsfig} 
\usepackage{mathptmx} 
\usepackage{times} 
\usepackage{amsmath} 
\usepackage{amssymb}  
\usepackage{graphicx}
\usepackage{epstopdf}
\usepackage{pdfpages}
\usepackage{cite}
\usepackage{amsbsy}
\usepackage{bm}
\usepackage{accents}
\newcommand*{\dt}[1]{%
  \accentset{\mbox{\bfseries .}}{#1}}
\newcommand*{\ddt}[1]{%
  \accentset{\mbox{\bfseries .\hspace{-0.25ex}.}}{#1}}
\pdfoutput=1
\usepackage{hyperref}
\hypersetup{
    colorlinks=true,
    linkcolor=black,
    citecolor=black,
    filecolor=black,
    urlcolor=black,
}

\def\horizontaldistance{\kern2pt}
\def\verticaldistance{5pt}

\makeatletter
\newcommand*{\rom}[1]{\expandafter\@slowromancap\romannumeral #1@}
\makeatother
\title{\LARGE \bf
Robust Bipedal Locomotion Control Based on  Model Predictive Control and Divergent Component of Motion
}

\author{Milad Shafiee-Ashtiani,$^{1}$ Aghil Yousefi-Koma,$^{1}$ and Masoud Shariat-Panahi$^{2}$  
\thanks{$^{1}$Center of Advanced Systems and Technologies (CAST)
School of Mechanical Engineering, College of Engineering, University of Tehran, Tehran, Iran.
       ( {\tt\small shafiee.a@ut.ac.ir}) ( {\tt\small aykoma@ut.ac.ir})}%
\thanks{$^{2}$School of Mechanical Engineering, College of Engineering, University of Tehran, Tehran, Iran.
       ( {\tt\small mshariatp@ut.ac.ir})}%
}

\begin{document}

\maketitle
\thispagestyle{empty}
\pagestyle{empty}

\begin{abstract}
In this paper, previous works on the Model Predictive Control (MPC) and the Divergent Component of Motion (DCM) for bipedal walking control are extended. To this end, we employ a single MPC which uses a combination of Center of Pressure (CoP) manipulation, step adjustment, and Centroidal Moment Pivot (CMP) modulation to design a robust walking controller. Furthermore, we exploit the concept of time-varying DCM to generalize our walking controller for walking in uneven surfaces. Using our scheme, a general and robust walking controller is designed which can be implemented on robots with different control authorities, for walking on various environments, e.g. uneven terrains or surfaces with a very limited feasible area for stepping. The effectiveness of the proposed approach is verified through simulations on different scenarios and comparison to the state of the art.
\end{abstract}

\section{INTRODUCTION}
In order to realize the dream of employing humanoid robots in our real world, developing a unified, robust and versatile framework for bipedal locomotion control is essential. That is why after several decades of research on bipedal locomotion, developing a robust and human-like walking controller is still one of the most  challenging  areas of the humanoid robotics research. Experimental studies show that the response of a human to progressively increasing disturbances can be categorized into three basic strategies: (1) ankle, (2) hip  and (3) stepping strategy and human uses these strategies  in a complex and efficient way consistent with the environment constraints . As a result, developing a walking controller that mimics the human behavior in different situations can significantly increase the reliability of these robots performing in a complex environment. This problem is the main goal of this paper.

Exploiting the whole dynamics of a humanoid robot is one of the common approaches that is used for walking trajectory generation \cite{orin2013centroidal}. However, solving a high dimensional nonlinear optimization problem demands high computation burden. As a result, simplified linear models that capture the task-relevant dynamics to a set of linear equations are useful for generating walking patterns in real-time. In this context, the Linear Inverted Pendulum Model (LIPM) \cite{kajita20013d}, has been very successfully used for the design of walking controllers for complex biped robots. 

Using the LIPM, Kajita \cite{kajita2003biped} introduced preview control (PC) method as an efficient tool for walking pattern generation based on the LIPM. Wieber  \cite{wieber2006trajectory} improved the robustness of this method by expressing PC as an MPC problem by taking into account the inequality constraints. It has been shown that the MPC based walking has a strong potential for disturbance rejection by exploiting the step location as a control input \cite{herdt2010online}. Similarly, the MPC  has been deployed for Push recovery using stepping strategy \cite{stephens2010push} \cite{aftab2012ankle}.

 On the other side, Pratt et al. \cite{pratt2006capture ,koolen2012capturability} introduced the Capture Point (CP) by splitting the COM dynamics into stable and unstable parts. Also, the CP has been used by Hof et al. \cite{hof2008extrapolated} to explain human walking properties under the name of extrapolated Center of Mass (XCoM). Takaneka et al. \cite{takenaka2009real} constrained the divergent part of the CoM to generate the DCM trajectory. Englsberger et al. \cite{englsberger2015three} extended the CP dynamics to the three-dimensional DCM, and developed a real-time walking controller. They also used the angular momentum in their CP controller \cite{englsberger2012integration}, but it has not been used in DCM trajectory generation. Furthermore, the Time-Varying DCM has been introduced for better tracking of the vertical component of the DCM during walking on uneven terrains\cite{hopkins2015dynamic}.

The idea of using the DCM dynamics instead of the whole CoM dynamics in an MPC framework has been proposed in \cite{krause2012stabilization}. However, they only used the CoP as the control input, while the DCM yields information about the step locations \cite{khadiv2016stepping}. To achieve more robust DCM based walking, Griffin et al. used the DCM concept through an MPC framework, and also considered the step locations as control inputs \cite{griffin2016model}. Moreover, Khadiv et al. showed that step timing adjustment play a key role in push recovery using step adjustment \cite{khadiv2016step}. However, to the best of our knowledge, the angular momentum, despite its high potential for disturbance rejection, has never been used in the DCM-based trajectory generation methods. The  whole-body angular momentum plays a key role in implementing robust walking \cite{lack2015integrating,herr2008angular}, particularly in the case where step adjustment is not possible. In the presence of significant disturbances, the ZMP approaches the safe margin and therefore  generating angular momentum or step adjustment is required for maintaining the balance \cite{lack2015integrating,popovic2005ground}. However, in the case where the step locations are limited, employing the centroidal angular momentum becomes crucial \cite{shafiee2016push,wiedebach2016walking}.

In this paper, inspired by above-mentioned works, we propose a novel walking controller with two main contributions. First, we develop a unified, versatile and robust walking controller capable of rejecting severe pushes, using the CoP manipulation, step adjustment, and CMP modulation simultaneously in a single MPC. This results in a significant improvement in terms of disturbance rejection. In the situation where the step adjustment is not possible, the significance of the CMP modulation or employing the centroidal angular momentum becomes more evident. Second, this method allows considering the change of the CoM height during DCM trajectory generation for walking on uneven terrains. The proposed method for changing the DCM height improves the  reverse-time integration method \cite{griffin2016model}, since using reverse-time integration may cause discontinuities in the DCM  trajectory in the case where more than one previewed step is required.

The rest of this paper is organized as follows. In Section \rom{2} we briefly review the LIPM dynamics, the MPC and DCM formulations. The proposed walking controller is presented in Section \rom{3}. In \rom{4}, the obtained simulations results are presented and discussed. Finally, Section \rom{5} concludes the findings.

\section{Center of Mass Dynamics}
\subsection{Linear Inverted Pendulum}

The LIPM  has already been used successfully  for describing the Center of Mass (CoM) dynamics for bipedal locomotion. (\cite{kajita20013d}-\cite{griffin2016model} ). Briefly, the LIPM uses the following assumptions:
\begin{itemize}

\item The rate change of centroidal angular momentum is zero,
\item The CoM height remains constant at $\Delta z$
\item The torque of base joint of the pendulum is zero
\end{itemize}
By these assumptions, the equations of Motion of the LIPM may be specified as:
\begin{equation}
\ddot x_{com}= \omega_{0}^{2}( x_{com}-p_{x}) , \quad \ddot y_{com}= \omega_{0}^{2}( y_{com}-p_{y}) 
 \label{eq:1}
\end{equation}

in which $x_{com}$ and $y_{com}$ are the CoM horizontal components,  $\omega_{0}=\sqrt{\frac{g}{\Delta z}}$ is the natural frequency of the pendulum,  and $p_{x}$,  $p_{y}$ are the horizontal components  of the CoP. 
 
The whole-body angular momentum as well as the change of CoM height play a key role for versatile and robust walking pattern and are ignored in the LIPM dynamics. These effects will be discussed in this paper.

\begin{figure}[]
\centering
      \mbox{\parbox{2in}{
\includegraphics[scale=.35, trim ={8.5cm 8.5cm 4.8cm 14.1cm},clip]{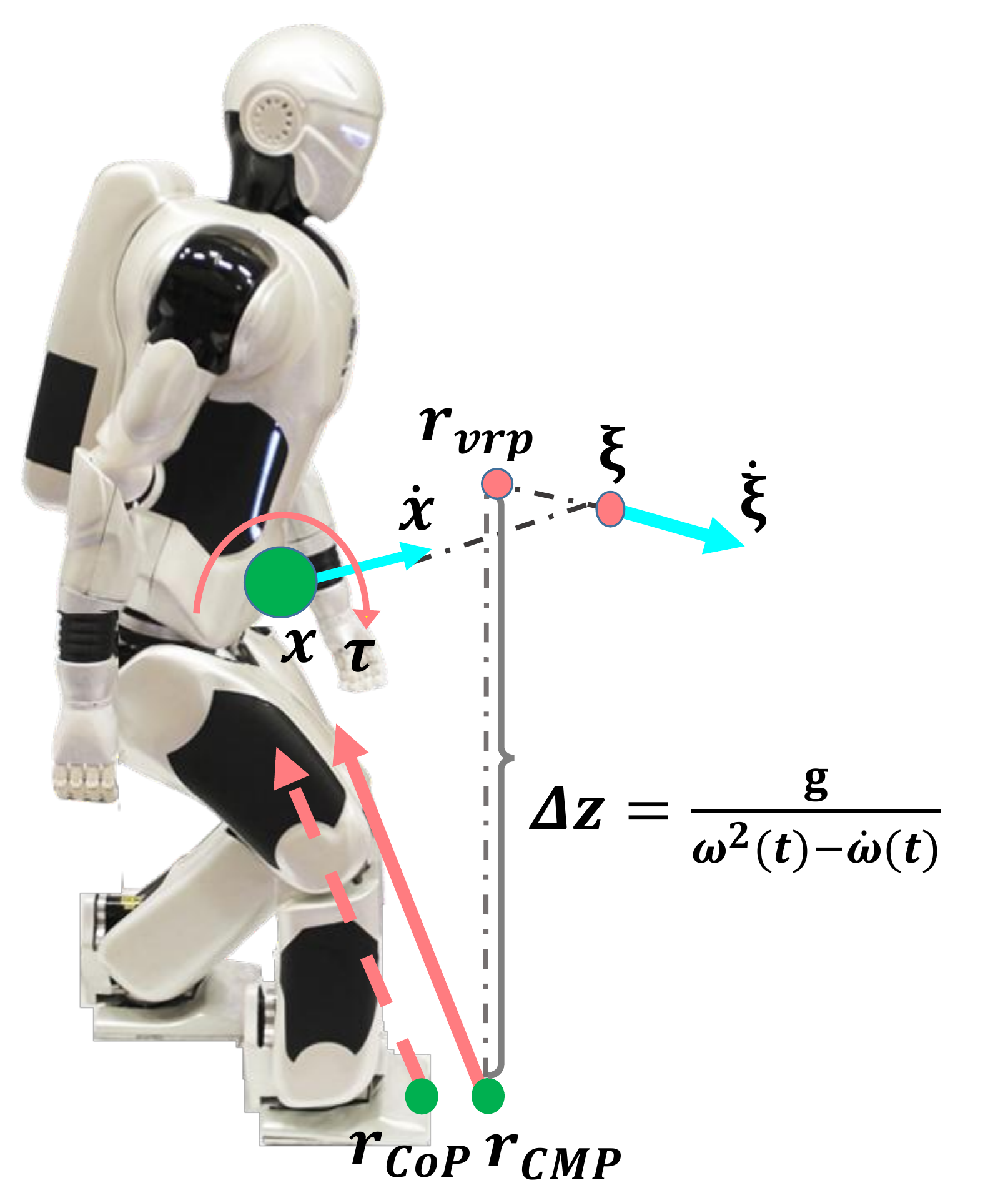}
}}

\caption{ DCM dynamics for SURENA \rom{3} humanoid robot during walking}
      \label{fig1}
    \end{figure}    
 \subsection{Time-Varying DCM}
The CoM dynamics can be split into its stable and unstable parts.  The unstable part is called DCM and is extended to the time-varying DCM by considering the change of the natural frequency of the LIPM. The 3D DCM is defined as: 
\begin{equation}
\boldsymbol{\xi}=\mathbf{ x}+\frac{\mathbf{\dt x}}{{\omega(t)}} 
  \label{eq:2}
\end{equation}
where $\mathbf{ x}=[ x_{com}, y_{com}, z_{com}]^T$ is the CoM position, $\mathbf{\dot x}$ is the CoM velocity, and $\omega(t) $ is the time-varying natural frequency.

From (\ref{eq:2}), the CoM dynamics is given by:
\begin{equation}
\mathbf{\dot x}={{\omega(t)}} (\boldsymbol{\xi}-\mathbf{ x} )
  \label{eq:3}
\end{equation}
assuming ${\omega(t)>\alpha}$ ($\alpha$ is a positive and small value), the Com dynamics can be shown to be asymptotically stable and it follows the DCM. By differentiating (\ref{eq:2}) and substituting (\ref{eq:3}) and defining $\mathbf{r}_{vrp}$ as the time-varying virtual repellent point (VRP) , the DCM dynamics is expressed as  :
\begin{equation}
\dt{\mathbf{ \xi}}=\Bigg({{\omega-\frac{\dt {\omega}}{\omega}}}\Bigg)(\boldsymbol{\xi}-\mathbf{r}_{vrp})
  \label{eq:4}
\end{equation}

Using the concept of the Centroidal Moment Pivot (CMP)  \cite{englsberger2015three}, the $\mathbf{r}_{vrp}$ can be related as a function of CMP :
\begin{equation}
\mathbf{r}_{vrp}=\mathbf{x}-\frac{\mathbf{\ddot x}}{{\omega}^2-\dt{\omega}}=\mathbf{r}_{cmp}+\frac{\mathbf{g}}{ {\omega}^2-\dt {\omega}}
  \label{eq:5}
\end{equation}

where   $\mathbf{g}=[ 0, 0, -g]^T$ is the gravity acceleration vector and $\mathbf{r}_{CMP}$ is defined as:
\begin{equation}
\mathbf{r}_{cmp}=\mathbf{x}-\frac{\mathbf{F}_{ext}}{{\omega}^2-\dt{\omega}}
  \label{eq:6}
\end{equation}

In this equation, $\mathbf{F}_{ext}$ stands for  the vector of contact forces. For generating DCM trajectory, $\mathbf{r}_{cmp}$  is assumed to coincide with  $\mathbf{r}_{cmp}$ in the ground plane. The CMP is equal to the CoP in the case of zero moment around the CoM. For a non-zero moment about the CoM, however, the CMP can move beyond the edges of the support polygon, while the CoP still remains inside the support polygon. In other words, the centroidal momentum pivot is the point where a line parallel to the ground reaction force and passing through the CoM intersects the ground. Therefore the relation between the CoP and the CMP is defined as:
\begin{equation}
\mathbf{r}_{cmp}=\mathbf{r}_{cop}+\frac{1}{ m({g}+{\ddot{z}})}\big[\dt{H}_y,-\dt{H}_x,0\big]^T
  \label{eq:7}
\end{equation}

 Where ${H}$ is the angular momentum around the CoM. 

\section{DCM based Model Predictive Controller }
In this section, we employ the MPC framework and the DCM dynamics to come up with a unified and robust framework for walking control of a biped robot. Englsberger et al. \cite{englsberger2015three} introduced the 3D DCM and designed a versatile and fast method for walking trajectory generation by having a predefined ZMP and a final condition on the DCM that should coincide with the final CoP. In \cite{griffin2016model}, the DCM acceleration and step positions are selected as control inputs for step adjustment, similar to \cite{herdt2010online}.The human walking analysis shows that the whole-body angular momentum is small and is highly regulated throughout the walking cycle \cite{herr2008angular}. In fact, human uses the effect of centroidal angular momentum in the situation where external disturbances are exerted, particularly in cases where the step adjustment is not possible. In order to consider these cases, we optimize angular momentum when an external disturbance acts on the robot. In fact, we define the problem as a single MPC which decides the step locations, the second derivative  of centroidal angular momentum, and the rate of change of CoP.

\subsection{DCM Discrete dynamics }
The discrete-time DCM dynamics using (\ref{eq:3}),(\ref{eq:5})and (\ref{eq:7}) can be specified as follows:
\begin{equation}
{\psi}_{k+1}={A}_{k}{\psi}_{k}+{B}_{k}{u}_{k}
  \label{eq:8}
\end{equation}

while the corresponding matrices in this equation are:
\[{\psi}_{k+1}=
\begin{bmatrix}
   \xi_{ k+1}  \\
    x_{k+1}      \\
   \dt{H}_{k+1}  \\
   {cop}_{k+1}   
\end{bmatrix}
 \horizontaldistance  \horizontaldistance
B_{k}=
\begin{bmatrix}
    0 &0  \\
    0 & 0  \\
    T & 0  \\
    0 & T 
\end{bmatrix} \horizontaldistance \horizontaldistance
{u}_{k}=
\begin{bmatrix}

   \ddt{H} _k \\
 \dt {cop}_k    \\
\end{bmatrix}
\]
 \[ 
A_{k}=\begin{bmatrix}
   T\frac{\omega ^2_{k}-\dt{\omega}_{k}}{\omega_{k}}+1  & 0 &-T\frac{\omega ^2_{k}-\dt{\omega}_{k}}{\omega_{k} (mg+m\ddot{ z})}  &  -T\frac{\omega ^2_{k}-\dt{\omega}_{k}}{\omega_{k}}\\[\verticaldistance]
\omega_{k} T  & (1-\omega_{k} T)  & 0 & 0 \\[\verticaldistance]
  0 &0 &1  &0  \\[\verticaldistance]
    0 &0 &0  &1  
\end{bmatrix}
\]
\\
Given a sequence of control inputs $u_k$, the linear model in (\ref{eq:8}) can be converted into a sequence of states $\boldsymbol\Gamma_{\psi}$, for a previewed number of time-steps ($N$):
\begin{equation}
\begin{aligned}
\boldsymbol{\Gamma_{\psi}}=\boldsymbol{\Phi}_{k}{\psi}_{k}+\boldsymbol{\Phi}_{u1}\boldsymbol{\Gamma_{u1}}\horizontaldistance  
 \label{eq:9}
\end{aligned}
\end{equation}
\begin{align*}
\boldsymbol{\Gamma_{\psi}}=[{\psi}_{k+1}^T  {\psi}_{k+2} ^T \horizontaldistance \horizontaldistance  \dots \horizontaldistance {\psi}_{k+N}^T]^T
\end{align*}
\begin{align*}
\boldsymbol{\Gamma_{u1}}=[{u}_{k}^T \horizontaldistance {u}_{k+1} ^T   \dots \horizontaldistance {u}_{k+N-1}^T]^T  
\end{align*}
$
\boldsymbol{\Phi}_{k}=
\begin{bmatrix}
   A_{ k}  \\
     A_{ k} \horizontaldistance A_{k+1}     \\
   \vdots  \\
    A_{ k} .. \horizontaldistance A_{k+N}   
\end{bmatrix}
 \horizontaldistance
\boldsymbol{\Phi}_{u1}=
\begin{bmatrix}
 B & 0 & ..& 0 \\
  A_{ k} B&B  & ..&0 \\
    A_{ k} \horizontaldistance A_{k+1}      \\
   \vdots &  \vdots & \ddots & \vdots  \\
      A_{ k} .. \horizontaldistance A_{k+N-1}B & .. &.. &   B
\end{bmatrix}
$
\\

It is noteworthy that the trajectory of $\dt\omega$ in each time step is given by knowing the vertical trajectory of the CoM during walking. We can use this equation for both lateral and sagittal directions, since the dynamics equations in the sagittal and lateral planes are decoupled.

\subsection{MPC Cost Function}  
After discretization of the dynamics equations, we can write the problem through a discrete-time MPC. The proposed MPC objective function is as follows:
\begin{equation}
\begin{aligned}
J= \sum_{\mathclap{k=1}}^{N} \alpha_{1}\|  \dt {cop}_{k}   \|^2 + \alpha_{2}\|  \dt H_{k+1}    \|^2 +  \alpha_{3}\|   \ddt H_{k}   \|^2+ \\  \alpha_{4}\|  cop_{k+1}-cop^{ref}_{k+1}    \|^2  +\alpha_{5}\|  \xi_{k+1}^{ref}- \xi_{k+1}   \|^2
\\[10pt]
\end{aligned}
 \label{eq:10} 
\end{equation}
Where $N$ is the number of time intervals and $\alpha_{i}$'s are the weights. The first term is used for minimization of the rate change of CoP  that helps to smoothing the contact forces to generate CoM smooth motion. The second term is presented for modulation CMP in the case of large disturbance by optimizing angular momentum. The third term was used for manipulating angular momentum and it's weight selected carefully to to smoothing rate change of angular momentum in situations where disturbances exist.  A theoretical analysis  of MPC shows that minimizing any derivative of the motion of the CoM of the robot while enforcing the constraints on the position of the CoP results in stable biped locomotion. Therefore the latter  term in (\ref{eq:10}) is used to maintaining a position  of the DCM as close as possible to some reference positions  for guiding the robot through desired direction and position in the environment particularly in the situation that the step adjustment is activated. Also this term act like a spring and damper to the profile of the CoM, results to better handling disturbances. 
\subsection{Automatic Step Adjustment}
In order to express the footstep positions as a control input of MPC, we have to express the position of the footholds over the previewed horizon  $\boldsymbol{\Gamma_{u2}}  \in \mathbb{R}^{ m}$ with the current given footsteps $\boldsymbol{\Gamma_{0}} \in \mathbb{R}^{N}$ which is fixed on the ground. The $ \boldsymbol{\Phi}_{0} \in \mathbb{R}^{N }$ and $ \boldsymbol{\Phi}_{u2} \in \mathbb{R}^{N \times m}$ contains 0 and 1 specify which time-steps $T_{i} $ belong to  which steps and $m$ is the number of previewed steps, so we have\cite{herdt2010online}:
\begin{equation}
\boldsymbol{\Gamma_{CoP}}=\boldsymbol{\Phi}_{0}\boldsymbol{\Gamma_{0}}+\boldsymbol{\Phi}_{u2}\boldsymbol{\Gamma_{u2}}
  \label{eq:11}
\end{equation}
\[\boldsymbol{\Phi}_{0}=
\begin{bmatrix}
  1  \\
    \vdots      \\
   1  \\
   0\\
 \vdots\\
0\\
0\\
\vdots\\
0\\
0\\
 \vdots\\
\vdots\\
0   
\end{bmatrix}
\boldsymbol{\Phi}_{u2}=\begin{bmatrix}
0 & 0  & & 0  \\
    \vdots &\ddots&& \vdots     \\
   0& 0&&0 \\
   1&  0 && 0\\
 \vdots &\ddots  & &\vdots\\
1& 0& &0    \\
0&1& &0  \\
 \vdots&\ddots  & & \vdots\\
0  &  1& \ddots & 0\\
0  & 0 &  &  \vdots \\

\vdots  & \vdots &  &\vdots \\
0   &0 &  & 1 
\end{bmatrix}
 \horizontaldistance  \horizontaldistance
\boldsymbol{\Gamma_{u2}}=
\begin{bmatrix}
p_{1}\\
p_{2}\\
 \vdots\\
 \vdots\\
p_{m}
\end{bmatrix}
\]

Therfore based on  (\ref{eq:9}) and (\ref{eq:11}) for the whole prediction horizon we have:
\begin{equation}
\begin{bmatrix}
 \boldsymbol{\Gamma_{\psi}}\\\\
\boldsymbol{\Gamma_{CoP}}
\end{bmatrix}
=
\begin{bmatrix}
\boldsymbol{\Phi}_{k}& 0\\
0 &   \boldsymbol{\Phi}_{0}
\end{bmatrix}
\begin{bmatrix}
{\psi}_{k}\\\\
\boldsymbol{\Gamma_{0}}
\end{bmatrix}
+
\begin{bmatrix}
\boldsymbol{\Phi}_{u1}&
0\\
0&\boldsymbol{\Phi}_{u2}
  \label{eq:12}
\end{bmatrix}
\begin{bmatrix}
   \boldsymbol{\Gamma_{u1}}\\\\
 \boldsymbol{\Gamma_{u2}}
\end{bmatrix}
\end{equation}
This equation is derived for $x$ direction however for the $y$ direction is the same. By using (\ref{eq:12}), the objective function of MPC can be expressed as a Quadratic Programming (QP) problem with constraints in the standard form of:
\begin{equation}
\begin{aligned}
J=\frac{1}2 \hspace{.1cm} {\boldsymbol\Gamma}_{ U}^T\hspace{.1cm} H \hspace{.1cm} {\boldsymbol\Gamma}_{U} +\hspace{.1cm} {\boldsymbol\Gamma}_{ U}^T\hspace{.1cm} f\\
st. \hspace{1.7cm}\\
C \hspace{.1cm} {\boldsymbol\Gamma}_{U}+D=0 \hspace{1.1cm}\\
E\hspace{.1cm} {\boldsymbol\Gamma}_{U}+F \leq0  \hspace{1.1cm}
\end{aligned}
  \label{eq:13}
\end{equation}
Where $C, D, E $ and $F$ are the coefficient matrices, $\boldsymbol \Gamma_{U} \in \mathbb{R}^{2N+m}$ is input vector for forward direction,  with $H$ and $f$ being the Hessian matrix and the gradient vector of the objective function.
\subsection{Constraints}
The main advantage of MPC method is consideration of future constraints. In all previous works on DCM trajectory generation,  the final condition on the DCM that should be coincided with the CoM and ZMP is used to produce the DCM boundary conditions.  Here we do this intelligently  by enforcing the final position constraint on the  DCM. It prevents the reverse-time integration that may cause discontinuities for the lack of initial condition and also enable a straightforward DCM trajectory generation method. At the end, we use a constraint to ensure the rate of angular momentum is zero at the final time step.
\begin{equation}
\begin{aligned}
\mathbf{cop_{N}}=\boldsymbol{\xi}_{N}\hspace{1.1cm}\\
\mathbf{cop_{N}}=\mathbf{x}_{N}\hspace{1.1cm}\\
\mathbf{\dt{H}_{N}}=0 \hspace{1.1cm}\\
 \mathbf{CoP}   \in   Support Polygon 
\end{aligned}
  \label{eq:14}
\end{equation}
The first three constraints are equality constraints for the last step time of implementation of QP. The latter is inequality constraint that is considered for overturn avoidance. Also for the step adjustment problem, bound on the position of the
next foothold simply enforced with the horizontal position of the CoM $\boldsymbol\Gamma_{x}$ as follows:
\begin{equation}
\begin{aligned}
 \|\boldsymbol\Gamma_{u2} -\boldsymbol\Gamma_{x} \| < l 
\end{aligned}
  \label{eq:15}
\end{equation}

\section{Results And Discussion}
In this section, we will show that the proposed algorithm can generate a robust walking locomotion in different scenarios. QP is solved using Gurobi solver by MATLAB software. Physical characteristic of our SURENA \rom{3} humanoid robot, with the desired CoM height of  $0.75  \hspace{0.05cm}cm $, and weight of $90  \hspace{0.05cm}kg $ which is developed at CAST is used for trajectory generation. Each step is split to a primary  double support phase (DSP), single support phase (SSP) and a secondary DSP. The primary DSP for the first step and secondary DSP for the last step are called initial and final DSP respectively.  These DSP have longer time than to middle steps DSP duration for smooth accelerating and decelerating of the CoM at the start and stop of the walking.
\subsection{Simulation results}
\begin{figure}[]
\centering
      \mbox{\parbox{4in}{
   \includegraphics[scale=.66, trim ={3.94cm 12.5cm 5.2cm 7.8cm},clip]{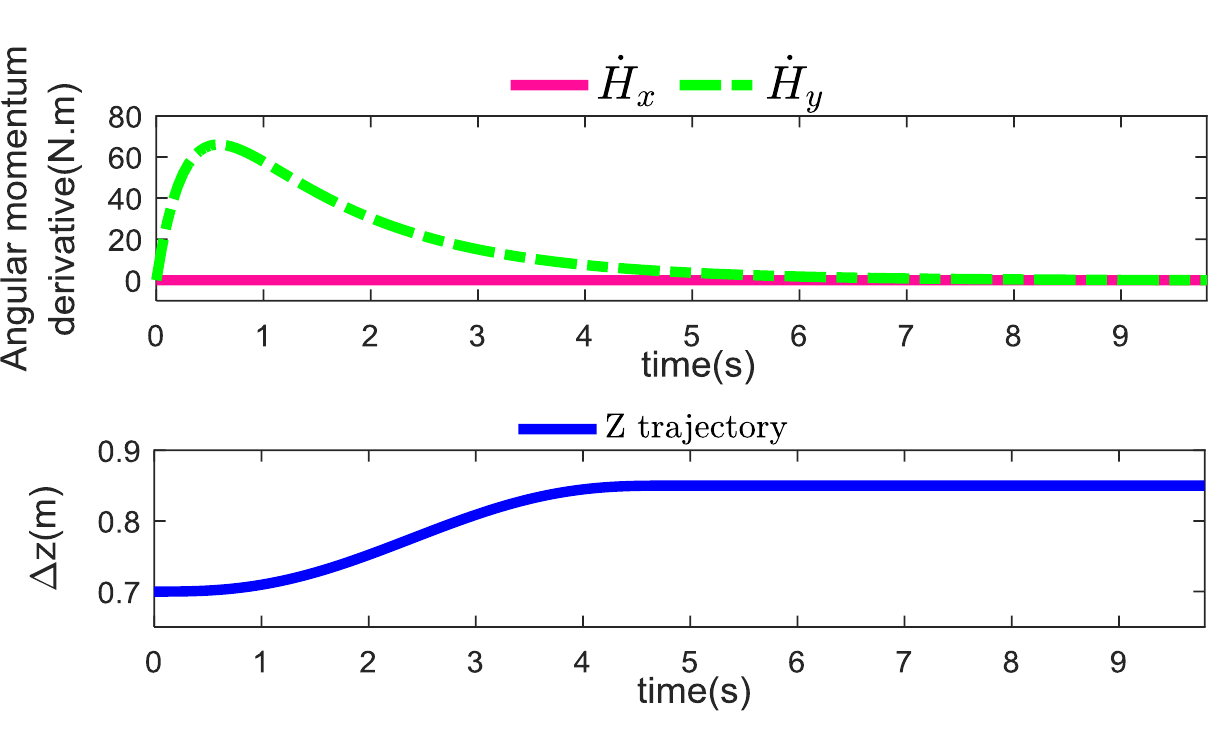}
  \includegraphics[scale=.730, trim ={4.94cm 10.5cm 5.5cm 8.3cm},clip]{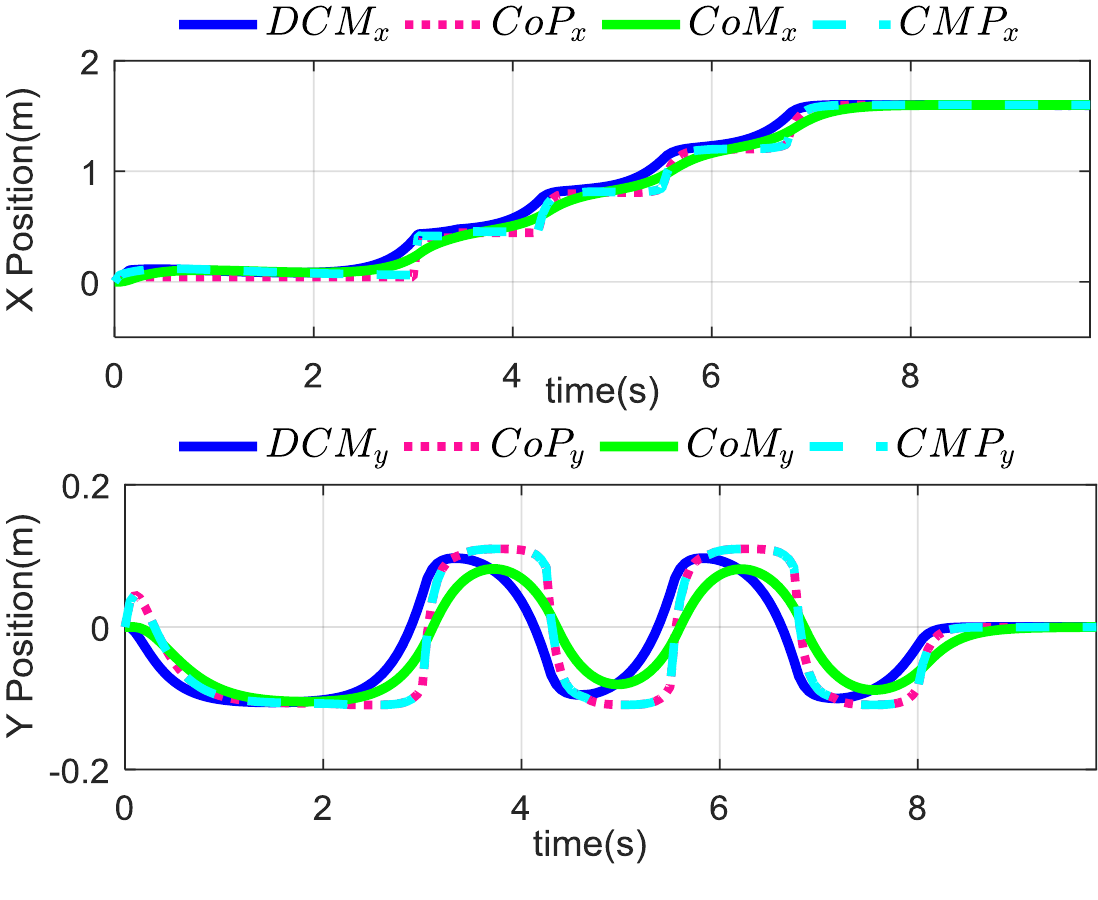}
 \includegraphics[scale=0.68, trim ={3.95cm 13.5cm 3cm 9cm},clip]{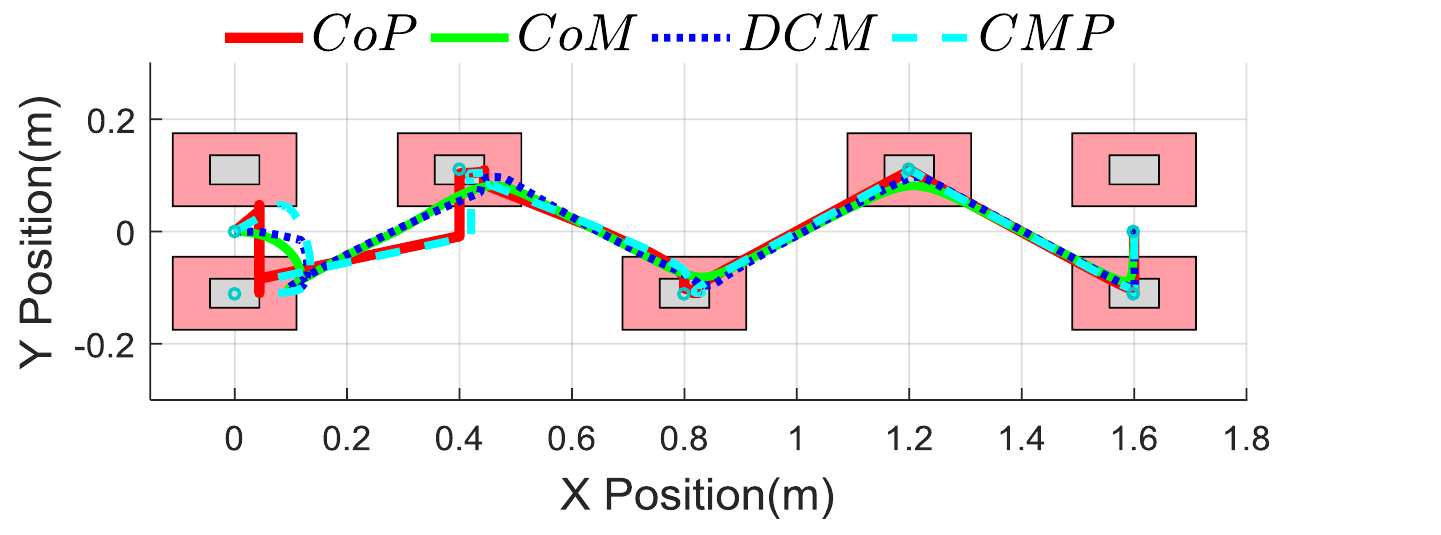}
}}
\caption{CMP modulation without step adjustment for 5 steps walking on limited contact surfaces with initial DSP duration of $2.16 \hspace{0.05cm}s$, final DSP duration of $1.8 \hspace{0.05cm}s$, middle DSP duration of $0.18 \hspace{0.05cm}s$ and single support phase (SSP) duration of $0.84  \hspace{0.05cm}s$ with step length of $0.4  \hspace{0.05cm}m$ with CoM height change of $15 \hspace{0.05cm}cm$, $250 \hspace{0.05cm}N$ forward push on the first step }
      \label{fig2}
    \end{figure}

 Fig. \ref{fig2} shows the result of trajectory generation for walking on limited contact surfaces. These plots show a simulation of 5 steps walking with initial DSP Duration of  $2.16  \hspace{0.05cm}s $, final DSP Duration of  $1.8 \hspace{0.05cm}s $, middle DSP duration of  $0.18 \hspace{0.05cm}s $ and single support phase (SSP) duration of  $0.84 \hspace{0.05cm}s $ with step length of  $0.4  \hspace{0.05cm}m $. In this simulation the previewed horizon is 5 steps and the part of step placement control of MPC is deactivated in order to track the exact desired footsteps. The inner gray rectangular shows the contact surface and the light pink rectangular shows the foot of robot.  In this scenario, a push with the magnitude of  $250 \hspace{0.05cm} N $
during $0.05 \hspace{0.05cm} s$ in sagittal direction is exerted on the CoM of the robot.
As we expected, the large push throws the DCM out of the
support polygon, and the CoP cannot navigate it and holds on the margin of support polygon. Therefore,
the angular momentum is generated by the MPC to move the
CMP outside the support polygon for controlling the DCM.  The results show that this MPC scheme can handle the change of CoM height in DCM trajectory generation. However, in this paper, our emphasis is not on walking on uneven terrain but we aim to show that the proposed controller is able to consider the change of natural frequency of DCM in straightforward way without reverse-time integration. Also Fig. \ref{fig2} shows that change of centroidal angular momentum is $70  \hspace{0.05cm}N.m$ that should be handled by whole-body motion of the robot. The advantage of this method for DCM trajectory generation is that it can handle the change of CoM height and centroidal angular momentum trajectory. Also for considering double support phase we do not require a different method and all aspects are integrated into a unique MPC scheme. This simulation shows the advantage of our MPC framework that enable to generate robust walking motion in the presence of large disturbances without changing the step position in situations that footstep must be placed on exact position for example when robot walks on rock and moreover contact surface is limited.

\subsection{Comparison with \cite{griffin2016model,herdt2010online}}
\begin{figure}[]
\centering
      \mbox{\parbox{4in}{
\includegraphics[scale=.69, trim ={4.2cm 11.9cm 3cm 9.2cm},clip]{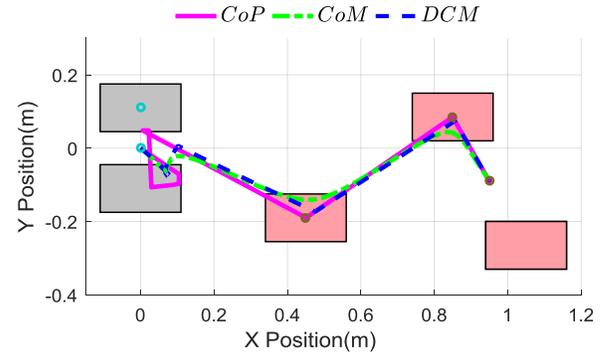}
}}
\caption{Step adjustment based on\cite{griffin2016model,herdt2010online}  for 3 steps walking with initial DSP duration of $1 \hspace{0.05cm}s$, final  DSP duration of $2 \hspace{0.05cm}s$, middle DSP duration of $0.2 \hspace{0.05cm}s$ and  SSP duration of $0.6 \hspace{0.05cm}s$, $100 \hspace{0.05cm}N$ lateral and $120 \hspace{0.05cm}N$ forward push on the first step.}
      \label{fig4}
    \end{figure}

\begin{figure}[]
\centering
      \mbox{\parbox{4in}{
   \includegraphics[scale=.72, trim ={4.3cm 11.9cm 3cm 9.8cm},clip]{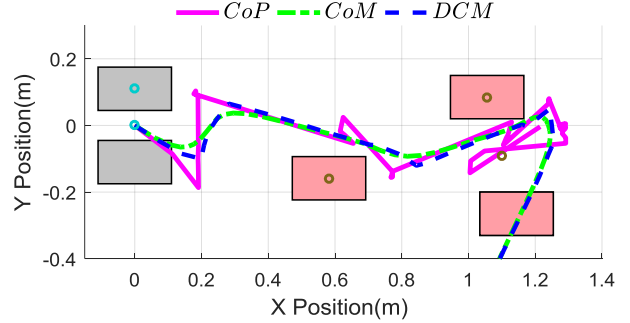}
}}
\caption{Diverging of MPC problem based on  \cite{griffin2016model,herdt2010online}  for  3 steps walking with initial DSP duration of $1 \hspace{0.05cm}s$, final  DSP duration of $2 \hspace{0.05cm}s$, middle DSP duration of $0.2 \hspace{0.05cm}s$ and SSP duration of $0.6 \hspace{0.05cm}s$, $200 \hspace{0.05cm}N$ lateral and $140 \hspace{0.05cm}N$ forward push on the first step.}
      \label{fig5}
    \end{figure}

\begin{figure}[]
\centering
      \mbox{\parbox{4in}{
   \includegraphics[scale=.68, trim ={4.3cm 12.2cm 3cm 8.2cm},clip]{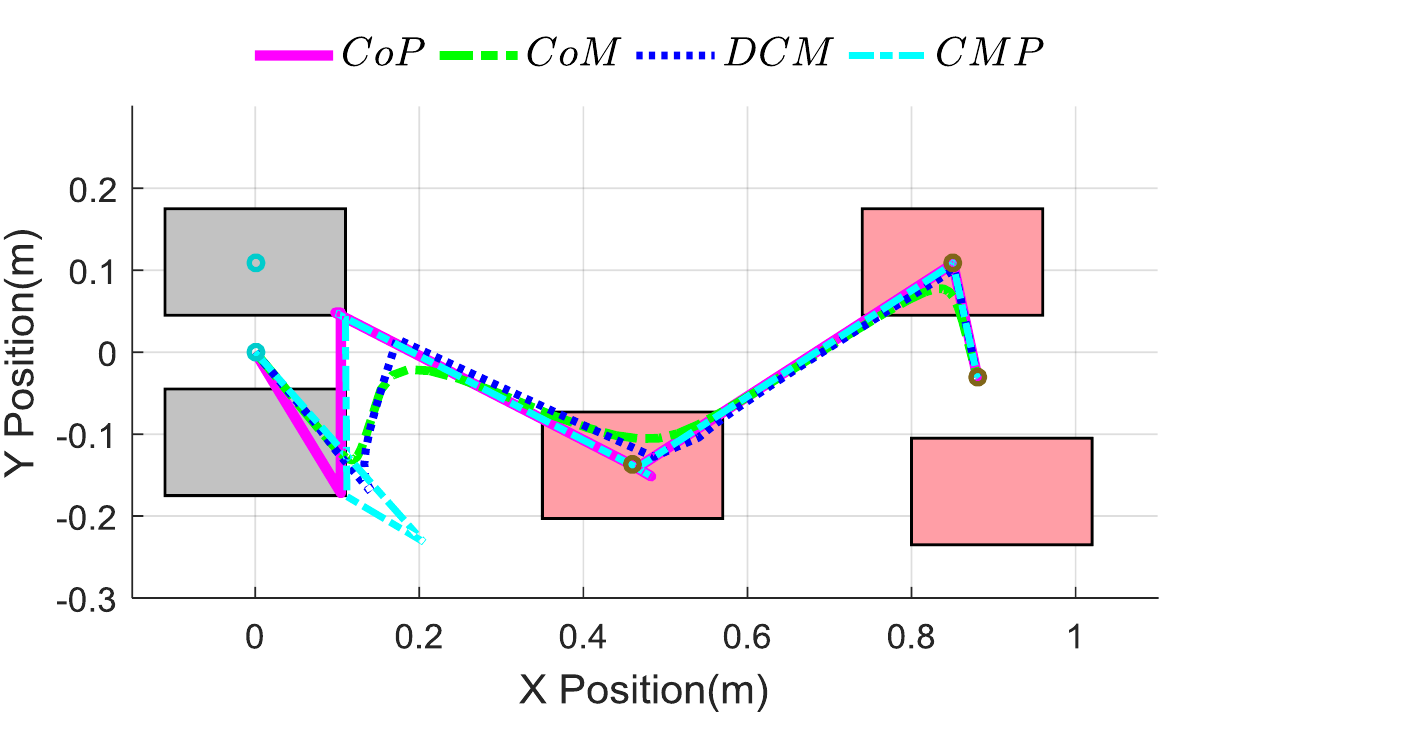}
\includegraphics[scale=.75, trim ={5.4cm 15.7cm 2.9cm 8.2cm},clip]{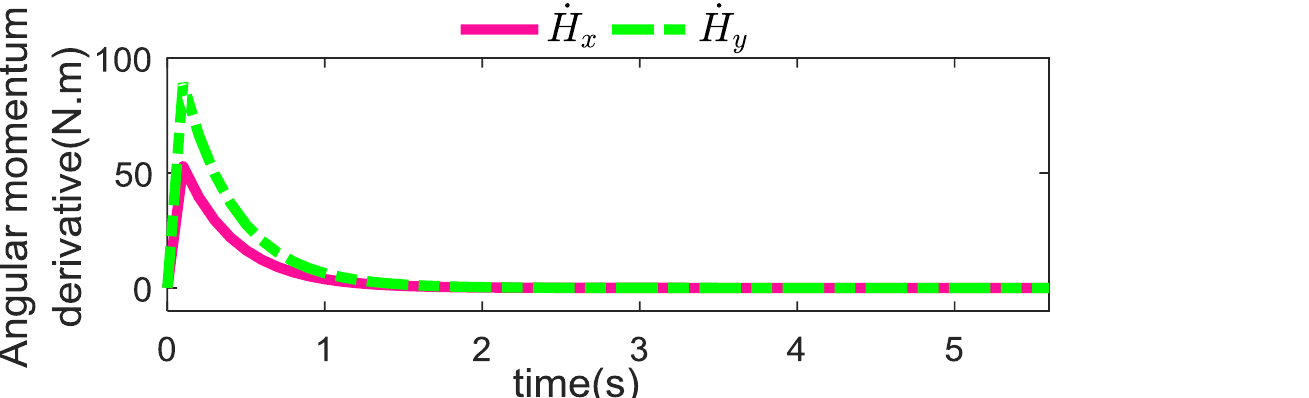}
}}
\caption{Step adjustment, CMP and CoP modulation based on proposed method  for 3 steps walking with initial DSP duration of $1 \hspace{0.05cm}s$ final  DSP duration of $2 \hspace{0.05cm}s$, middle DSP duration of $0.2 \hspace{0.05cm}s$ and SSP duration of $0.6 \hspace{0.05cm}s$ $260 \hspace{0.05cm}N$ lateral and $220 \hspace{0.05cm}N$ forward push on the first step}
      \label{fig6}
    \end{figure}

In the second scenario, we compare the robustness of our proposed optimization procedure with step adjustment to the proposed approach in \cite{griffin2016model,herdt2010online} that is the cutting edge of  walking locomotion planners. The algorithm that is presented in \cite{griffin2016model} is extension of (\cite{herdt2010online, englsberger2015three,hopkins2015dynamic}) which  are standard walking pattern generators in the literature.  That is why we compare our results to this approach. We applied the same parameters for both approaches using an LIPM and computed the maximum push that each approach can handle with characteristic of SURENA \rom{3} humanoid robot. For each simulation, uniformly forces during $ \Delta \hspace{0.05cm} t = 0.1 \hspace{0.05cm}s$ are applied at the first step. The previewed horizon is 3 steps and the height of CoM fixed at $85\hspace{0.05cm} cm$.
As it shown in Fig. \ref{fig4}  the algorithm of \cite{griffin2016model,herdt2010online} can handle the pushes with magnitude of $100 \hspace{0.05cm} N$ lateral and $120 \hspace{0.05cm} N$ using automatic step placement. As  it can be observed in Fig.\ref{fig5} this approach fails to recover pushes with magnitude more than  $140 \hspace{0.05 cm}N$ lateral and  $200 \hspace{0.05cm} N$ forward. Our approach with CoP  and CMP modulation can recover  larger  severe pushes compared to the approach in \cite{griffin2016model,herdt2010online}. Also our approach needs shorter step length in comparison with \cite{griffin2016model,herdt2010online} for step adjustment. Fig. \ref{fig6} shows that CMP is outside of support polygon by generating $90 \hspace{0.05cm} N.m $ rate of centroidal angular momentum but CoP holds on the edge of  the foot. The MPC scheme of \cite{griffin2016model,herdt2010online} causes that the forward and outward step length increased up to bound of constraint $5.5  \hspace{0.05cm}cm$, $0.1  \hspace{0.05cm}cm$ respectively,  that  causes actuator saturation. However our method can handle significant larger disturbance with step length adjustment of $4.8 \hspace{0.05cm} cm$ forward and $0.05  \hspace{0.05cm}cm$ outward using CMP modulation.
Based on obtained results the proposed method has the following advantages to other methods:
\begin{itemize}
\item  In our proposed approach, we employed the step location and CoP and CMP modulation for locomotion control in a single MPC that results better robustness than conventional one, in simulation.  
\item  Our experiments suggested that our approach can consider the change of  CoM height. The proposed  controller can compensate the severe pushes, when the robot walks on small
contact surfaces  such as rock, also is capable
of saving the robot from falling in the situations that
step adjustment is not possible.
\end{itemize}

\section{Conclusions}
In this paper, we proposed a method for generating robust walking locomotion for biped robots using the concept of time-varying DCM by employing the MPC. In this method, a combination of CoP manipulation, step adjustment, and CMP modulation is exploited through a single MPC to generate robust and human-like walking patterns for a previewed number of steps. Using the idea of time-varying DCM, our controller is capable of dealing with uneven terrains. However, in this paper, our emphasis is developing robust walking control  and the walking on uneven terrain will be discussed in the future works. The results from simulation on different scenarios showed the robustness of our controller in various situations. Also,  Comparison to the state of the art shows significant improvement in disturbance rejection capabilities of  DCM-based walking controllers.
Despite all above advantages, this controller is implemented
only in simulation. Implementing on the real robot has more practical challenges such as
accurate state estimation to obtain the DCM position, the saturation
of actuators especially in the case where the support
polygon is small. Generating whole-body motion of the robot for the required angular momentum is an another challenge that can be handled if the robot body equipped with a momentum gyro! Our future works will be considering change of step timing in the MPC scheme to achieve more robust walking motion and experimental implementation on a real robot.




\bibliographystyle{IEEEtran}
\bibliography{Master}

\begin{thebibliography}{10}
\providecommand{\url}[1]{#1}
\csname url@samestyle\endcsname
\providecommand{\newblock}{\relax}
\providecommand{\bibinfo}[2]{#2}
\providecommand{\BIBentrySTDinterwordspacing}{\spaceskip=0pt\relax}
\providecommand{\BIBentryALTinterwordstretchfactor}{4}
\providecommand{\BIBentryALTinterwordspacing}{\spaceskip=\fontdimen2\font plus
\BIBentryALTinterwordstretchfactor\fontdimen3\font minus
  \fontdimen4\font\relax}
\providecommand{\BIBforeignlanguage}[2]{{%
\expandafter\ifx\csname l@#1\endcsname\relax
\typeout{** WARNING: IEEEtran.bst: No hyphenation pattern has been}%
\typeout{** loaded for the language `#1'. Using the pattern for}%
\typeout{** the default language instead.}%
\else
\language=\csname l@#1\endcsname
\fi
#2}}
\providecommand{\BIBdecl}{\relax}
\BIBdecl

\bibitem{orin2013centroidal}
D.~E. Orin, A.~Goswami, and S.-H. Lee, ``Centroidal dynamics of a humanoid
  robot,'' \emph{Autonomous Robots}, vol.~35, no. 2-3, pp. 161--176, 2013.

\bibitem{kajita20013d}
S.~Kajita, F.~Kanehiro, K.~Kaneko, K.~Yokoi, and H.~Hirukawa, ``The 3d linear
  inverted pendulum mode: A simple modeling for a biped walking pattern
  generation,'' in \emph{Intelligent Robots and Systems, 2001. Proceedings.
  2001 IEEE/RSJ International Conference on}, vol.~1.\hskip 1em plus 0.5em
  minus 0.4em\relax IEEE, 2001, pp. 239--246.

\bibitem{kajita2003biped}
S.~Kajita, F.~Kanehiro, K.~Kaneko, K.~Fujiwara, K.~Harada, K.~Yokoi, and
  H.~Hirukawa, ``Biped walking pattern generation by using preview control of
  zero-moment point,'' in \emph{Robotics and Automation, 2003. Proceedings.
  ICRA'03. IEEE International Conference on}, vol.~2.\hskip 1em plus 0.5em
  minus 0.4em\relax IEEE, 2003, pp. 1620--1626.

\bibitem{wieber2006trajectory}
P.-B. Wieber, ``Trajectory free linear model predictive control for stable
  walking in the presence of strong perturbations,'' in \emph{2006 6th IEEE-RAS
  International Conference on Humanoid Robots}.\hskip 1em plus 0.5em minus
  0.4em\relax IEEE, 2006, pp. 137--142.

\bibitem{herdt2010online}
A.~Herdt, H.~Diedam, P.-B. Wieber, D.~Dimitrov, K.~Mombaur, and M.~Diehl,
  ``Online walking motion generation with automatic footstep placement,''
  \emph{Advanced Robotics}, vol.~24, no. 5-6, pp. 719--737, 2010.

\bibitem{stephens2010push}
B.~J. Stephens and C.~G. Atkeson, ``Push recovery by stepping for humanoid
  robots with force controlled joints,'' in \emph{2010 10th IEEE-RAS
  International Conference on Humanoid Robots}.\hskip 1em plus 0.5em minus
  0.4em\relax IEEE, 2010, pp. 52--59.

\bibitem{aftab2012ankle}
Z.~Aftab, T.~Robert, and P.-B. Wieber, ``Ankle, hip and stepping strategies for
  humanoid balance recovery with a single model predictive control scheme,'' in
  \emph{2012 12th IEEE-RAS International Conference on Humanoid Robots
  (Humanoids 2012)}.\hskip 1em plus 0.5em minus 0.4em\relax IEEE, 2012, pp.
  159--164.

\bibitem{pratt2006capture}
J.~Pratt, J.~Carff, S.~Drakunov, and A.~Goswami, ``Capture point: A step toward
  humanoid push recovery,'' in \emph{2006 6th IEEE-RAS international conference
  on humanoid robots}.\hskip 1em plus 0.5em minus 0.4em\relax IEEE, 2006, pp.
  200--207.

\bibitem{koolen2012capturability}
T.~Koolen, T.~De~Boer, J.~Rebula, A.~Goswami, and J.~Pratt,
  ``Capturability-based analysis and control of legged locomotion, part 1:
  Theory and application to three simple gait models,'' \emph{The International
  Journal of Robotics Research}, vol.~31, no.~9, pp. 1094--1113, 2012.

\bibitem{hof2008extrapolated}
A.~L. Hof, ``The ‘extrapolated center of mass’ concept suggests a simple
  control of balance in walking,'' \emph{Human movement science}, vol.~27,
  no.~1, pp. 112--125, 2008.

\bibitem{takenaka2009real}
T.~Takenaka, T.~Matsumoto, and T.~Yoshiike, ``Real time motion generation and
  control for biped robot-1 st report: Walking gait pattern generation,'' in
  \emph{2009 IEEE/RSJ International Conference on Intelligent Robots and
  Systems}.\hskip 1em plus 0.5em minus 0.4em\relax IEEE, 2009, pp. 1084--1091.

\bibitem{englsberger2015three}
J.~Englsberger, C.~Ott, and A.~Albu-Sch{\"a}ffer, ``Three-dimensional bipedal
  walking control based on divergent component of motion,'' \emph{IEEE
  Transactions on Robotics}, vol.~31, no.~2, pp. 355--368, 2015.

\bibitem{englsberger2012integration}
J.~Englsberger and C.~Ott, ``Integration of vertical com motion and angular
  momentum in an extended capture point tracking controller for bipedal
  walking,'' in \emph{2012 12th IEEE-RAS International Conference on Humanoid
  Robots (Humanoids 2012)}.\hskip 1em plus 0.5em minus 0.4em\relax IEEE, 2012,
  pp. 183--189.

\bibitem{hopkins2015dynamic}
M.~A. Hopkins, D.~W. Hong, and A.~Leonessa, ``Dynamic walking on uneven terrain
  using the time-varying divergent component of motion,'' \emph{International
  Journal of Humanoid Robotics}, vol.~12, no.~03, p. 1550027, 2015.

\bibitem{krause2012stabilization}
M.~Krause, J.~Englsberger, P.-B. Wieber, and C.~Ott, ``Stabilization of the
  capture point dynamics for bipedal walking based on model predictive
  control,'' \emph{IFAC Proceedings Volumes}, vol.~45, no.~22, pp. 165--171,
  2012.

\bibitem{khadiv2016stepping}
M.~Khadiv, S.~Kleff, A.~Herzog, S.~A. Moosavian, S.~Schaal, L.~Righetti
  \emph{et~al.}, ``Stepping stabilization using a combination of dcm tracking
  and step adjustment,'' \emph{in Robotics and Mechatronics (ICRoM), 2016 4th
  RSI/ISM International Conference on, Available: arXiv preprint
  arXiv:1609.09822}, 2016.

\bibitem{griffin2016model}
R.~J. Griffin and A.~Leonessa, ``Model predictive control for dynamic footstep
  adjustment using the divergent component of motion,'' in \emph{2016 IEEE
  International Conference on Robotics and Automation (ICRA)}.\hskip 1em plus
  0.5em minus 0.4em\relax IEEE, 2016, pp. 1763--1768.

\bibitem{khadiv2016step}
M.~Khadiv, A.~Herzog, S.~A.~A. Moosavian, and L.~Righetti, ``Step timing
  adjustment: A step toward generating robust gaits,'' in \emph{Humanoid Robots
  (Humanoids), 2016 IEEE-RAS 16th International Conference on}.\hskip 1em plus
  0.5em minus 0.4em\relax IEEE, 2016, pp. 35--42.

\bibitem{lack2015integrating}
J.~Lack, ``Integrating the effects of angular momentum and changing center of
  mass height in bipedal locomotion planning,'' in \emph{Humanoid Robots
  (Humanoids), 2015 IEEE-RAS 15th International Conference on}.\hskip 1em plus
  0.5em minus 0.4em\relax IEEE, 2015, pp. 651--656.

\bibitem{herr2008angular}
H.~Herr and M.~Popovic, ``Angular momentum in human walking,'' \emph{Journal of
  Experimental Biology}, vol. 211, no.~4, pp. 467--481, 2008.

\bibitem{popovic2005ground}
M.~B. Popovic, A.~Goswami, and H.~Herr, ``Ground reference points in legged
  locomotion: Definitions, biological trajectories and control implications,''
  \emph{The International Journal of Robotics Research}, vol.~24, no.~12, pp.
  1013--1032, 2005.

\bibitem{shafiee2016push}
M.~Shafiee-Ashtiani, A.~Yousefi-Koma, M.~Shariat-Panahi, and M.~Khadiv, ``Push
  recovery of a humanoid robot based on model predictive control and capture
  point,'' \emph{in Robotics and Mechatronics (ICRoM), 2016 4th RSI/ISM
  International Conference on, Available: arXiv preprint arXiv:1612.08034},
  2016.

\bibitem{wiedebach2016walking}
G.~Wiedebach, S.~Bertrand, T.~Wu, L.~Fiorio, S.~McCrory, R.~Griffin, F.~Nori,
  and J.~Pratt, ``Walking on partial footholds including line contacts with the
  humanoid robot atlas,'' in \emph{Humanoid Robots (Humanoids), 2016 IEEE-RAS
  16th International Conference on}.\hskip 1em plus 0.5em minus 0.4em\relax
  IEEE, 2016, pp. 1312--1319.

\end{thebibliography}

\end{document}